\pdfoutput=1
\documentclass{article}

\usepackage[preprint]{neurips_2025}
\makeatletter\renewcommand{\@noticestring}{}\makeatother
\usepackage{graphicx}
\usepackage{natbib}
\usepackage{amsmath}
\usepackage{booktabs}
\usepackage{tabularx}
\usepackage{array}
\usepackage{graphicx}
\usepackage{makecell}

\usepackage[utf8]{inputenc} 
\usepackage[T1]{fontenc}    
\usepackage{hyperref}       
\usepackage{url}            
\usepackage{booktabs}       
\usepackage{amsfonts}       
\usepackage{nicefrac}       
\usepackage{microtype}      
\usepackage{xcolor}         
\usepackage{ragged2e}       

\providecommand{\UseTaggingSocket}[1]{}
\ExplSyntaxOn
\cs_if_exist:NF \tbl_update_cell_data: { \cs_new:Npn \tbl_update_cell_data: {} }
\ExplSyntaxOff

\title{PolyGnosis 2.0: Enhancing LLM Reasoning via Agentic Harness Engineering for Polymarket and OSINT Insight Extraction}

\author{%
  Daren Wang \\
  The Chinese University of Hong Kong \\
  \texttt{darenwang@link.cuhk.edu.hk} \\
  \And
  Hong Xu \\
  The Chinese University of Hong Kong \\
  \texttt{hongxu@cuhk.edu.hk} \\
  \And
  Jiawen Xian \\
  Evolution AI Lab \\
  \texttt{xianjw@evolutionasset.com} \\
}

\begin{document}

\maketitle

\begin{abstract}
  This paper introduces \textbf{PolyGnosis 2.0}, a pioneering multi-agent architecture designed to extract predictive intelligence by synthesizing \textit{Polymarket} anomaly signals with global Open Source Intelligence (\textit{OSINT}) streams, specifically Global Database of Events, Language, and Tone (\textit{GDELT}). We define and target ``Perspective Mismatches'', the narrative divergence between Polymarket sentiment and global media flows, as high-alpha trading signals. Moving beyond generic agentic superiority, we rigorously quantify the efficacy of ``Harness Engineering'' techniques, including reflection loops, tool-calling, divide-and-conquer partitioning (D\&C), and chain-of-thought (CoT), within high-noise financial domains. Our empirical evaluation against human-expert benchmarks reveals that while structural partitioning is mandatory for multi-dimensional alignment, unconstrained terminal reflection actively induces logical drift. Furthermore, we identify a pervasive ``consensus bias'' across all agent configurations during narrative reasoning, necessitating deterministic validation. Ultimately, we isolate a Pareto-optimal configuration that achieves professional-grade analytical precision while minimizing latency and token overhead, providing a robust blueprint for autonomous intelligence in prediction markets.
\end{abstract}
\section{Introduction}
Prediction markets, notably \textit{Polymarket}\footnote{https://polymarket.com/}, serve as rich repositories of publicly observable, consensus-driven forecasting data. In these markets, topics are typically structured as binary propositions regarding future real-world events. Market prices, constrained between 0 and 1, function as collective probability estimations aggregated from participant behavior. Furthermore, the transparency of order books and detailed trader histories theoretically enables granular insight mining. However, the inherent complexity and noise of this data present a substantial barrier to systematic analysis. While Large Language Models (LLMs) offer a foundation for cross-domain data synthesis, their unaided internal reasoning may not consistently capture the intricate causalities required for professional-grade trading insights. Addressing these structural complexities need not rely exclusively on standard model scaling. Instead, a highly effective approach involves the strategic design of agentic frameworks, leveraging ``Harness Engineering'' to better elicit the structured, high-order reasoning necessary for such complex environments \citep{wu2024autogen, khattab2023dspy}.

To this end, this research introduces \textbf{PolyGnosis 2.0}, an AI Agent architecture designed to process massive anomaly datasets from Polymarket. The system systematically cross-validates these market signals against Open Source Intelligence (OSINT) streams. Specifically, it leverages the Global Database of Events, Language, and Tone (GDELT) \citep{leetaru2013gdelt} to execute this comparative analysis. Inherently, Polymarket signals reflect only the localized information and cognitive biases of its active trader pool. Relying solely on this endogenous data lacks the external context necessary to evaluate the true validity of market sentiment. Therefore, a comparative external information source is strictly required. By introducing GDELT as a comprehensive proxy for global mass media dynamics, our framework bridges this informational gap. Our primary objective is to generate professional-grade market insights by capturing the ``Perspective Mismatch'', the explicit narrative divergence between localized Polymarket traders and broader global media coverage. Such mismatches are widely recognized as critical trading signals in both quantitative and fundamental investment strategies.

Furthermore, this study rigorously validates the utility of ``Harness Engineering''. We investigate whether sophisticated multi-agent architectures, incorporating mechanisms such as reflection loops, tool-calling, Divide-and-Conquer (D\&C), Chain-of-Thought (CoT), can significantly amplify a system's reasoning capabilities without upgrading the underlying LLM base model. Ultimately, we demonstrate that this harnessing engineering can effectively substitute or augment the rigorous investment research processes traditionally conducted by human experts.

Our core contributions are summarized as follows:

\begin{itemize}
    \item \textbf{Dataset \& Benchmark Formulation:} We collect and formulate a comprehensive benchmark dataset comprising continuous Polymarket alerts, paired with ground-truth annotations generated by professional traders for both market clustering and trading insight analysis.
    
    \item \textbf{Agentic Harness Engineering Framework:} We design and implement a multi-agent system featuring Clustering, Keywords Extraction, and Analysis agents. This framework also mitigates single-pass LLM reasoning bottlenecks through several advanced harness designs.
    
    \item \textbf{Strictly Isolated Evaluation Methodology:} We design a comprehensive evaluation protocol with strict physical isolation to prevent temporal data leakage and prompt contamination. The system's performance is systematically quantified across four dimensions: accuracy, consistency, latency, and computational cost.
    
    \item \textbf{Empirical Validation of Harness Efficacy:} Our experiments demonstrate that with appropriate harness engineering, LLM-based agents can significantly overcome their baseline reasoning limitations. The proposed architecture achieves analytical performance closely approaching that of human professionals, while maintaining high consistency at a reasonable level of time and computation.
\end{itemize}

\section{Related works}
The integration of Artificial Intelligence within prediction markets represents an emerging research domain. Existing literature predominantly focuses on the direct application of Large Language Models (LLMs) or relies on domain-specific fine-tuning to process prediction market data \citep{yang2025llm, turtel2025outcome, turtel2025llms}. While these approaches establish baseline capabilities, the application of complex, multi-agent frameworks, specifically designed to systematically analyze prediction market dynamics through advanced harness engineering, remains a critical gap in current methodologies.

Concurrently, the deployment of LLMs for Open Source Intelligence (OSINT) analysis has gained significant traction. Due to their inherent proficiency in processing massive text-based information flows, LLMs are increasingly utilized within agentic architectures for automated intelligence gathering \citep{palmieri2025framework, berzinji2024utilisation, chen2026cyberthreat}. Extensive global databases such as GDELT are frequently paired with LLMs, primarily to construct dynamic knowledge graphs \citep{myers2025talking}. However, leveraging these OSINT streams as an external mechanism to cross-validate endogenous prediction market sentiment and identify narrative mismatches represents an underexamined intersection.

To address the limitations of single-pass model reasoning, recent computational advancements emphasize the development of agentic structures and scaffolding mechanisms. It is well-established that sophisticated frameworks can substantially amplify a model's reasoning capabilities without requiring structural modifications or scaling of the underlying foundation models \citep{wu2024autogen, khattab2023dspy, hong2023metagpt}. Prominent techniques central to this engineering approach include Chain-of-Thought (CoT) reasoning paradigms \citep{wei2022chain, yao2022react}, iterative reflection protocols \citep{shinn2023reflexion}, and automated tool-calling \citep{schick2023toolformer}. By synthesizing these components, our research applies rigorous harness engineering to systematically navigate and extract actionable insights from the complex data environments of prediction markets.

\section{Agent Architecture}

\subsection{Sequential Agentic Workflow}

\begin{figure}[htbp]
  \centering
  \includegraphics[width=\linewidth]{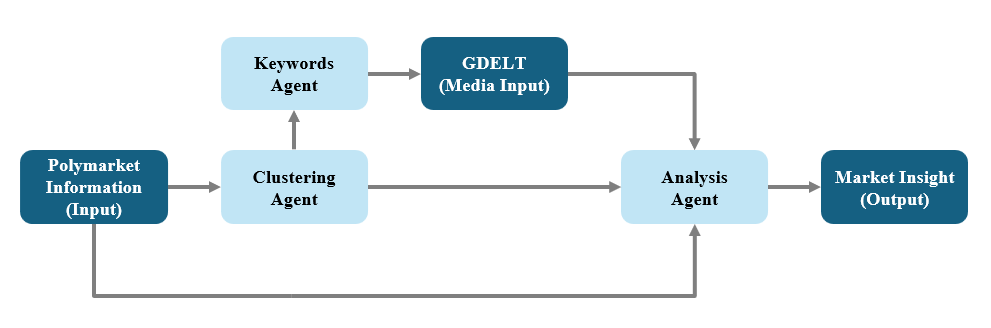}
  \caption{Sequential Agentic Workflow: Polymarket information enters the Clustering Agents; the Keywords Agent drives GDELT media retrieval; the Analysis Agent fuses GDELT signals, cluster structure, and raw Polymarket context into market insight.}
  \label{fig:workflow_sequential}
\end{figure}

The primary objective of the system is to systematically process anomaly signals from Polymarket, retrieve corresponding Open Source Intelligence (OSINT) via GDELT, and conduct comparative analysis to generate actionable insights. Engineered as a sequential workflow, the architecture integrates a non-agentic event-triggering module with three specialized intelligent agents. As illustrated in Figure~\ref{fig:workflow_sequential}, the end-to-end pipeline routes raw Polymarket inputs through sequential stages of clustering, keyword extraction, and media retrieval, concluding with a synthesis phase that evaluates market context and clustering data to generate insights through deep reasoning. The specific components of this phase are as follows:

\textbf{Trigger and Collection Module (Non-Agent):} This component continuously monitors Polymarket activity in real time via WebSockets. It generates automated alerts upon detecting predefined market anomalies, such as significant price volatility or large-volume transactions (whale activities). Concurrently, it aggregates metadata, including market propositions and historical trading behaviors, for identified whale accounts.

\textbf{Clustering Agent:} Functioning as a structural filter, this LLM-powered agent processes the aggregated market data to reduce noise. It executes thematic clustering, transforming unstructured event descriptions into standardized categories (e.g., Middle East Geopolitics, Cryptocurrency, U.S. Elections).

\textbf{Keywords Agent:} Operating on the categorized topics and specific event contexts provided by the Clustering Agent, this module extracts targeted search parameters. It interfaces directly with the GDELT database to retrieve relevant historical and real-time media coverage. Furthermore, it captures the sentiment scores and narrative trend metrics natively generated by GDELT's analytical engines.

\textbf{Analysis Agent:} As the terminal node in the workflow, this agent synthesizes the raw Polymarket signals, the structured cluster classifications, and the OSINT media intelligence. It applies multi-dimensional reasoning to produce a final, structured market insight. This output explicitly details the prevailing Polymarket trajectory, the dominant media narrative, the perspective mismatch between the two environments, and the underlying reasoning.

\subsection{Multi-Agent System with Enhanced Reasoning Harness}

\begin{figure}[htbp]
  \centering
  \includegraphics[width=\linewidth]{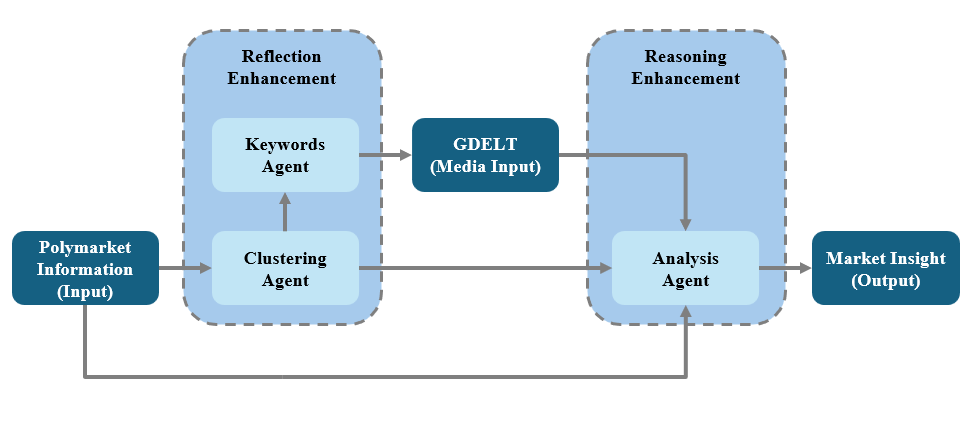}
  \caption{Multi-Agent System with Enhanced Reasoning Harness. Reflection Enhancement covers clustering and keyword--GDELT retrieval; Reasoning Enhancement centers on the Analysis Agent.}
  \label{fig:harness_enhanced}
\end{figure}

Since the foundational workflow involves complex reasoning tasks, relying on a standard, single-pass LLM pipeline may present operational constraints. To explore potential structural improvements, we introduce a Multi-Agent System architecture incorporating ``Reflection and Reasoning Enhancement'' harnesses, as illustrated in Figure~\ref{fig:harness_enhanced}.

\textbf{Reflection Enhancement:} The granularity of information clustering can influence the relevance of downstream news retrieval and subsequent analysis. Because a single-pass approach may face challenges in maintaining consistent clustering logic, we integrate agentic mechanisms, such as reflection loops, multi-agent collaboration, and tool calling. These additions are designed with the aim of aligning the classification process more closely with professional trading standards prior to media retrieval.

\textbf{Reasoning Enhancement:} Synthesizing multi-dimensional data (combining Polymarket information, structured clusters, and GDELT media inputs) requires robust logical deduction. To test whether the potential reasoning bottlenecks of single-generation LLM outputs can be mitigated, this module incorporates architectural patterns such as divide and conquer (D\&C), chain of thought (CoT), and reflection verification. This enhanced framework is hypothesized to facilitate deeper analysis and improve the reliability of the generated market insights.

\section{Experiment}

\subsection{Data Collection}
To rigorously evaluate the proposed system, we collected a dataset comprising Polymarket anomaly alerts, including price shocks and large-volume transactions, collected over five consecutive 24-hour UTC windows from March 25, 2026, 02:00 to March 30, 2026, 02:00. This raw alert stream serves as the foundational input for our empirical analysis. To establish a reliable Ground Truth (GT), professional traders manually annotated the dataset. Table~\ref{tab:exp1_data_collection} details the daily alert volume, the total number of thematic clusters identified, and the top five GT clusters by alert frequency within each evaluation window.

\begin{figure}[htbp]
  \centering
  \includegraphics[width=\linewidth]{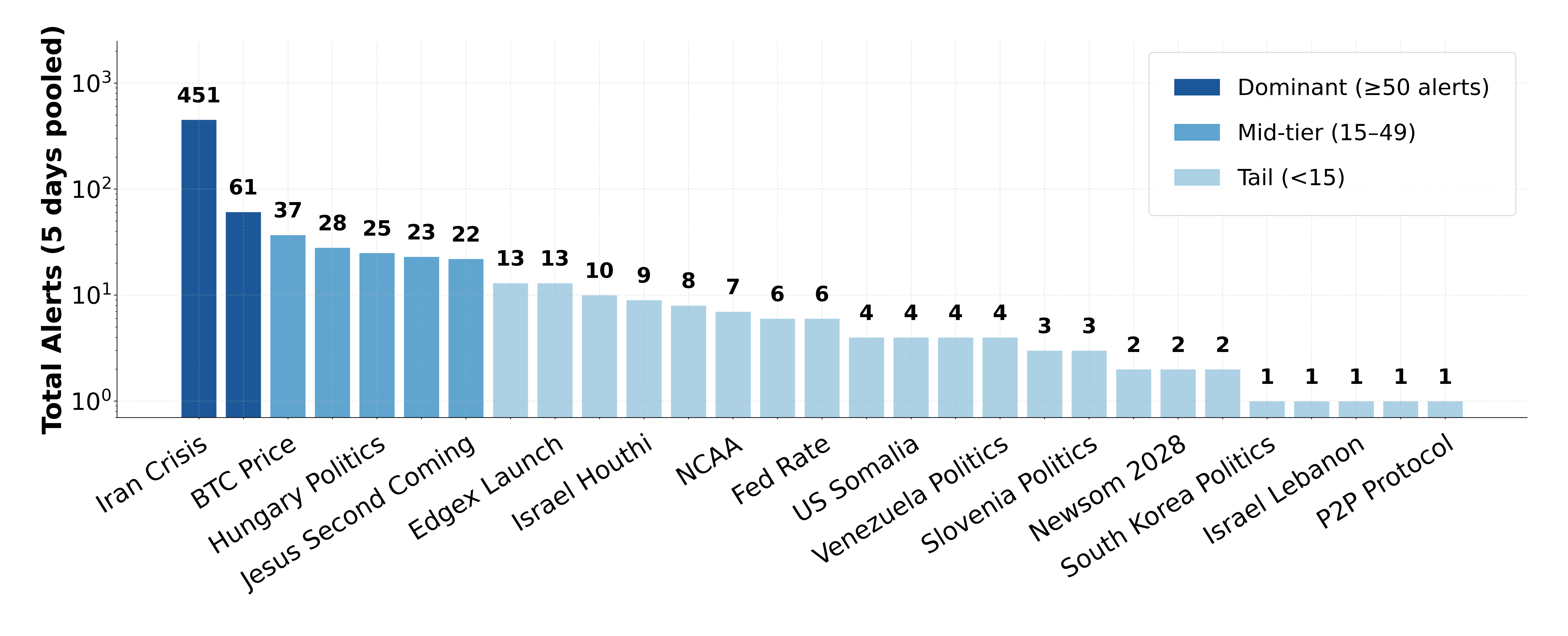}
  \caption{Alert volume by ground-truth cluster (five evaluation days pooled, logarithmic vertical axis), bars sorted by descending count. \emph{Note:} The dominant cluster (\texttt{Iran Crisis}) accounts for 451 of 752 total alerts ($\sim$60\%). The x-axis does not show every label to limit crowding.}
  \label{fig:cluster_longtail}
\end{figure}

During the annotation process, traders assigned each alert to a specific thematic category, thereby defining the GT for evaluating the Clustering Agent. As illustrated in Figure~\ref{fig:cluster_longtail}, the pooled cluster-alert distribution across the five-day period exhibits a pronounced long-tail characteristic. A concentrated subset of macro-themes (e.g., Iran Crisis, Crude Oil, BTC Price) dominates the alert volume, whereas the majority of categories register minimal activity. These human-assigned labels provide the baseline for quantifying the Clustering Agent's classification accuracy.

\begin{figure}[htbp]
  \centering
  \includegraphics[width=\linewidth]{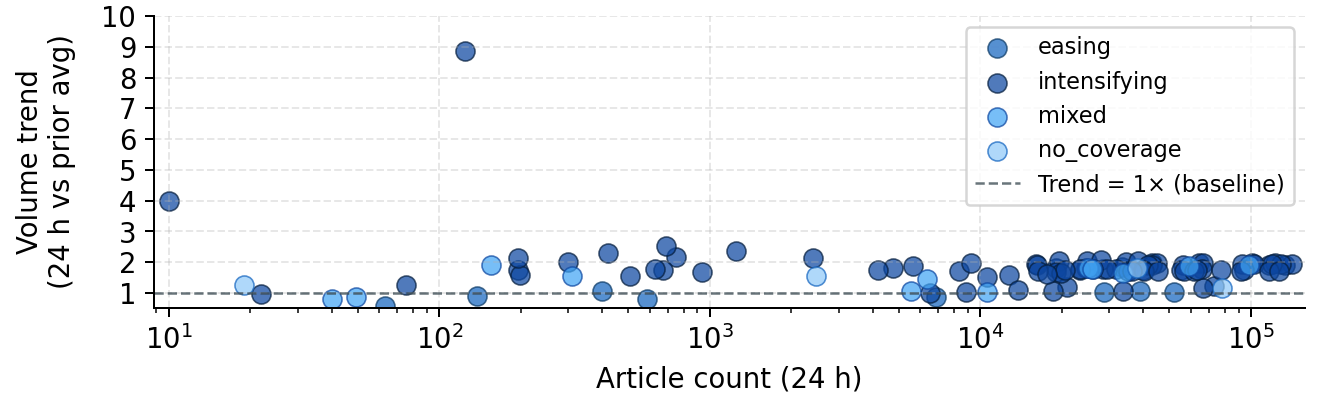}
  \caption{GDELT Global Knowledge Graph (GKG) media signal by cluster (24h article count vs.\ volume trend; colour encodes ground-truth media direction).}
  \label{fig:gdelt_coverage}
\end{figure}

To isolate downstream reasoning performance from potential upstream clustering errors, we held the cluster assignments constant during subsequent testing. The human-annotated GT clusters were fed directly into the pipeline to drive the retrieval of GDELT media signals and Polymarket context. Figure~\ref{fig:gdelt_coverage} visualizes the resulting GDELT Global Knowledge Graph (GKG) media coverage and media-direction annotations corresponding to these clusters.

\begin{figure}[htbp]
  \centering
  \includegraphics[width=\linewidth]{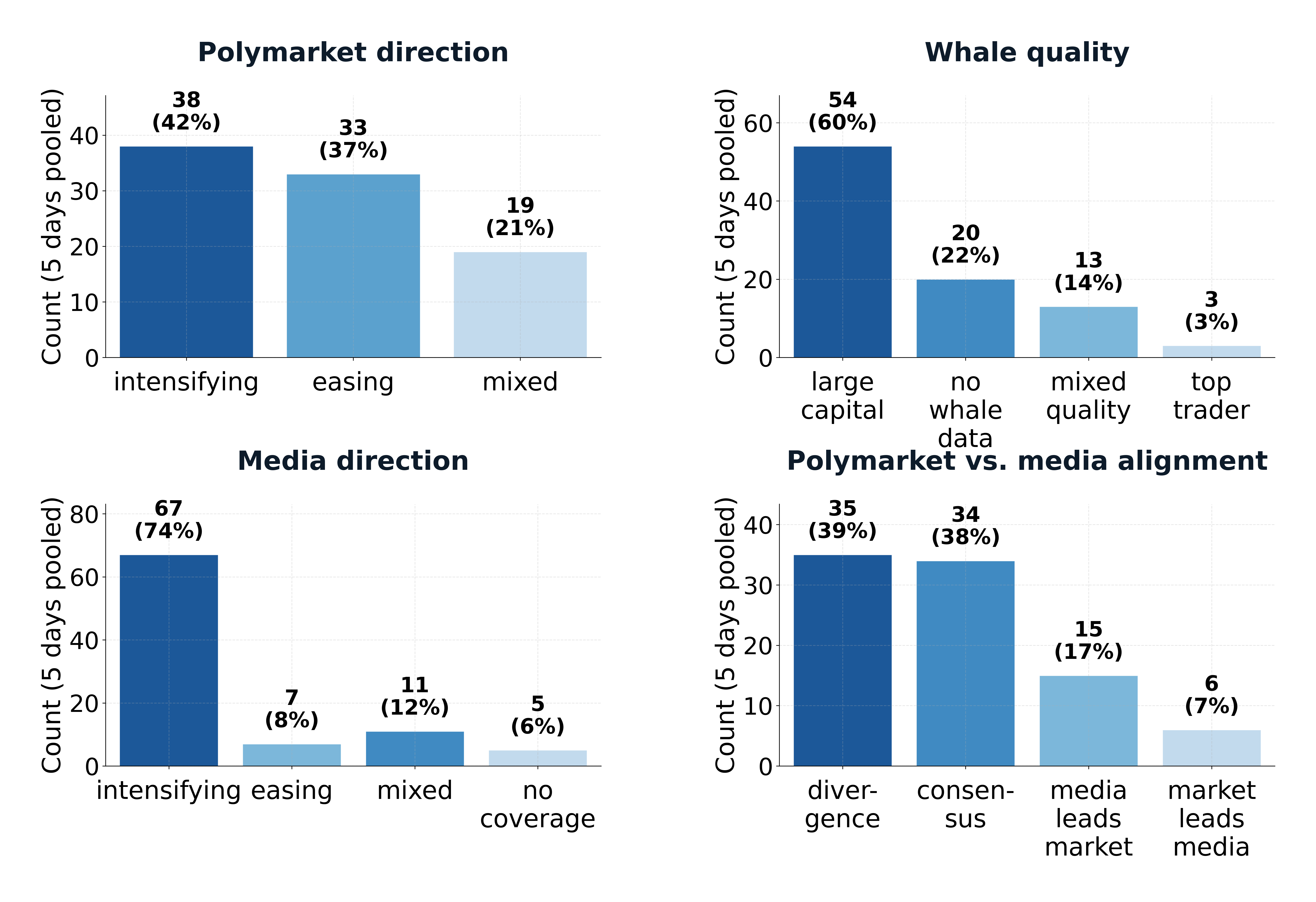}
  \caption{Ground truth label distributions for Analysis Agent. Frequency of human-assigned labels, pooled across five evaluation days and all clusters.}
  \label{fig:gt_label_dist}
\end{figure}

Recognizing the inherent complexity of extracting holistic market insights directly from raw, unstructured data streams, we structured the evaluation task into distinct analytical components. Expert traders were instructed to answer four discrete classification questions, which represent the critical intermediate reasoning steps required for synthesizing professional market insights. These questions entailed structured labeling of: (1) Polymarket direction, (2) whale-account quality, (3) media direction, and (4) Polymarket-media alignment. The frequency distribution of these labels across the pooled sample is presented in Figure~\ref{fig:gt_label_dist}. These expert choices constitute the formal GT used to systematically evaluate the reasoning and synthesis capabilities of the Analysis Agent.

\newcommand{\topfivecell}[1]{%
  \parbox[t]{\linewidth}{%
    \vspace{0pt}%
    \footnotesize\RaggedRight
    #1%
  }%
}

\begin{table}[htbp]
  \centering
  \caption{Polymarket alerts and human clustering (ground truth) statistics by day. Each column is one 24-hour window.}
  \label{tab:exp1_data_collection}
  \begingroup
  \small
  \setlength{\tabcolsep}{6pt}
  \renewcommand{\arraystretch}{1.12}
  \setlength{\extrarowheight}{2pt}
  \renewcommand{\tabularxcolumn}[1]{p{#1}}
  \begin{tabularx}{\textwidth}{@{} >{\raggedright\arraybackslash}p{2.15cm} *{5}{>{\RaggedRight\arraybackslash}X} @{}}
    \toprule
    \textbf{2026} & \textbf{Mar-25/26} & \textbf{Mar-26/27} & \textbf{Mar-27/28} & \textbf{Mar-28/29} & \textbf{Mar-29/30} \\
    \midrule
    \# Polymarket alerts & \multicolumn{1}{>{\centering\arraybackslash}X}{147} & \multicolumn{1}{>{\centering\arraybackslash}X}{159} & \multicolumn{1}{>{\centering\arraybackslash}X}{194} & \multicolumn{1}{>{\centering\arraybackslash}X}{104} & \multicolumn{1}{>{\centering\arraybackslash}X}{148} \\
    \# Ground truth clusters & \multicolumn{1}{>{\centering\arraybackslash}X}{15} & \multicolumn{1}{>{\centering\arraybackslash}X}{18} & \multicolumn{1}{>{\centering\arraybackslash}X}{22} & \multicolumn{1}{>{\centering\arraybackslash}X}{18} & \multicolumn{1}{>{\centering\arraybackslash}X}{17} \\
    \midrule
    Top-5 clusters (by count) &
    \topfivecell{Iran Crisis (81)\par Trump China Visit (14)\par Crude Oil (9)\par Trump Presidency (9)\par Jesus Second Coming (7)} &
    \topfivecell{Iran Crisis (80)\par Crude Oil (23)\par Trump Presidency (9)\par Trump China Visit (8)\par Hungary Politics (7)} &
    \topfivecell{Iran Crisis (114)\par Crude Oil (18)\par BTC Price (11)\par Hungary Politics (7)\par Edgex Launch (6)} &
    \topfivecell{Iran Crisis (77)\par BTC Price (4)\par Trump Presidency (3)\par Crude Oil (2)\par Netanyahu (2)} &
    \topfivecell{Iran Crisis (99)\par BTC Price (11)\par Crude Oil (9)\par Israel Houthi Hezbollah (5)\par Trump China Visit (3)} \\
    \bottomrule
  \end{tabularx}
  \endgroup
\end{table}

\subsection{Experimental Procedure}

We designed two parallel experimental frameworks to independently isolate and quantitatively assess the performance of the Clustering Agent and the Analysis Agent. Performance across both experiments was evaluated along four critical dimensions:

\begin{itemize}
    \item \textbf{Accuracy:} Alignment of model outputs with the human-annotated ground truth.
    \item \textbf{Consistency:} Variance and stability of the agent's output across multiple independent iterations.
    \item \textbf{Latency:} End-to-end system execution time.
    \item \textbf{Computational Cost:} Token usage and economic overhead.
\end{itemize}

To ensure ecological validity within simulated real-world trading environments, experiments were conducted under strict data isolation protocols. Prompt structures deliberately excluded any test-set information to prevent event specific few-shot contamination. Furthermore, the evaluation dataset's temporal window strictly succeeded the underlying Large Language Model's (LLM) knowledge cutoff date. Rigorous temporal boundaries were also enforced on all external search API calls to preclude the ingestion of future data, thereby systematically eliminating look-ahead bias.

\subsubsection{Experiment 1: Clustering Capability Evaluation}

This experiment utilized the Gemini 2.5 Flash model (knowledge cutoff as of Jan-25), selected for its computational efficiency in high-frequency classification tasks \citep{comanici2025gemini}. The evaluation comprised three distinct methodological tracks, each executing 10 independent runs over the five-day dataset:

\begin{itemize}
    \item \textbf{Direct Baseline:} The LLM executed classification without supplementary architectural scaffolding.
    \item \textbf{Reflection Loop:} A dual-agent reflection mechanism was integrated. An independent Reflection Agent iteratively reviewed and refined the Clustering Agent's outputs to optimize classification granularity. Specifically, this reflection protocol evaluated the semantic similarity among the cluster-specific keywords extracted by the Keywords Agent. Therefore, the system can systematically adjust and correct category boundaries. This iterative refinement process was bounded by a maximum of three cycles.
    \item \textbf{Reflection + Tool-Calling:} The Reflection Agent was augmented with external retrieval capabilities. By executing real-time queries via the Google Search API, the agent can reference news dynamics to better contextualize and ground its clustering decisions.
\end{itemize}

\subsubsection{Experiment 2: Reasoning and Analysis Capability Evaluation}

To evaluate complex logical reasoning and long-context synthesis, we deployed the Gemini 2.5 Pro model (knowledge cutoff as of Jan-25) \citep{comanici2025gemini} at the Analysis Agent level. This experiment featured four progressive architectural configurations:

\begin{itemize}
    \item \textbf{Global Context:} All thematic clusters related information were ingested simultaneously to assess the model's capacity for synthesizing global insights and identifying cross-cluster correlations.
    \item \textbf{Divide-and-Conquer (D\&C):} The dataset was partitioned by thematic cluster, and analysis was executed independently for each discrete category.
    \item \textbf{D\&C + CoT:} Building upon the D\&C track, the agent was explicitly prompted to generate a Chain of Thought (CoT), enforcing a transparent, step-by-step reasoning protocol.
    \item \textbf{D\&C + CoT + Reflection:} An independent Reflection Agent was introduced to the D\&C + CoT architecture to audit and correct the primary agent's logical deductions, constrained to a single verification iteration.
\end{itemize}

The primary quantitative metrics for Experiment 2 were classification accuracy and inter-run stability regarding the four ground-truth classification dimensions established by expert traders. Additionally, to evaluate the structural consistency of the free-text market insights, we utilized \textit{sentence-transformers} to compute the pairwise cosine semantic similarity across the independent runs, thereby quantifying the stability of the generative reasoning process.

\subsection{Results}

\subsubsection{Experiment 1: Clustering}
Table~\ref{tab:exp1_results} presents the quantitative outcomes of the clustering phase across the three architectural tracks. The empirical results demonstrate that augmenting the baseline LLM with a \textit{Reflection Loop} yields substantial improvements in clustering fidelity, increasing the Adjusted Rand Index (ARI) from $0.728 \pm 0.221$ to $0.883 \pm 0.142$ and Completeness from $0.786 \pm 0.119$ to $0.909 \pm 0.090$. The \textit{Direct Baseline} achieves the highest Homogeneity ($0.973 \pm 0.028$) but the lowest Completeness, revealing a systematic over-clustering problem. Without a revision step, the single-pass model is highly sensitive to prompt variability and divides ambiguous alerts into an excessive number of separate clusters (averaging $22.3$ per run compared to the ground truth average of $18$). 

\begin{figure}[htbp]
  \centering
  \includegraphics[width=\linewidth]{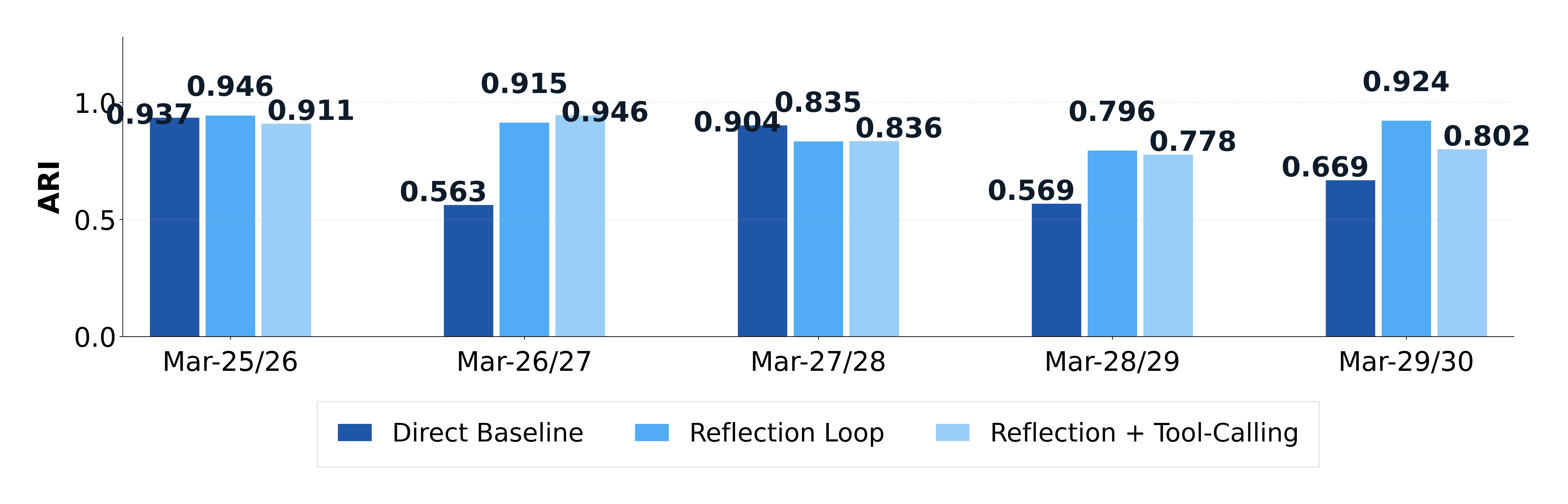}
  \caption{Experiment~1 Adjusted Rand Index (ARI) by day across the three clustering tracks (Direct Baseline, Reflection Loop, Reflection + Tool-Calling).}
  \label{fig:exp1_ari_by_day}
\end{figure}

\begin{figure*}[htbp]
  \centering
  \includegraphics[width=\textwidth]{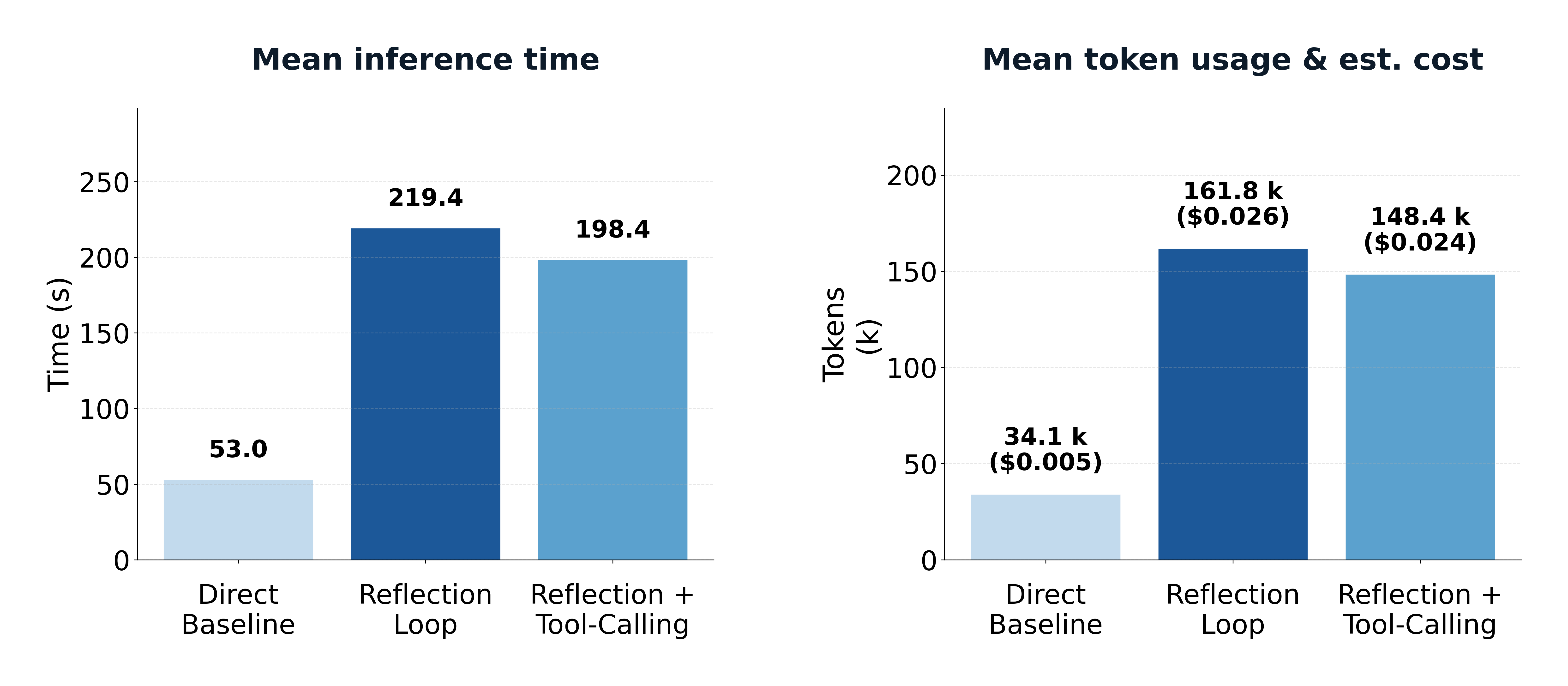}
  \caption{Experiment~1 mean inference time (left) and mean token usage with estimated cost (right) by track. Bars report track means aligned with Table~\ref{tab:exp1_results}.}
  \label{fig:exp1_time_tokens_cost}
\end{figure*}

The integration of external retrieval (\textit{Reflection + Tool-Calling}) maintains comparable overall performance metrics (ARI of $0.855 \pm 0.152$, Normalized Mutual Information of $0.906 \pm 0.063$) while slightly optimizing the validation process, reducing average reflection iterations from $3.00$ to $2.82 \pm 0.48$. However, this structural enhancement introduces a deliberate operational trade-off. As referenced in Figure~\ref{fig:exp1_time_tokens_cost}, the elevated classification accuracy of the reflection paradigms necessitates approximately a four-fold increase in latency (e.g., $219.4$ seconds vs. $53.0$ seconds) and token utilization relative to the \textit{Direct Baseline}, establishing a distinct cost-performance frontier for high-frequency market signal processing. For a detailed breakdown of daily ARI performance across the three tracks, see Figure~\ref{fig:exp1_ari_by_day}.

\begin{table}[htbp]
  \centering
  \caption{Experiment~1 clustering results (mean $\pm$ std over 50 runs, 5 days x 10 runs/day), ARI = Adjusted Rand Index; NMI = Normalized Mutual Information.}
  \label{tab:exp1_results}
  \begingroup
  \small
  \setlength{\tabcolsep}{8pt}
  \renewcommand{\arraystretch}{1.85}
  \setlength{\extrarowheight}{4pt}
  \renewcommand{\tabularxcolumn}[1]{m{#1}}
  \begin{tabularx}{\textwidth}{@{} >{\raggedright\arraybackslash}m{3.1cm} *{3}{>{\centering\arraybackslash}X} @{}}
    \toprule
    \textbf{Metric} &
    \makecell{\textbf{Direct}\\\textbf{Baseline}} &
    \makecell{\textbf{Reflection}\\\textbf{Loop}} &
    \makecell{\textbf{Reflection +}\\\textbf{Tool-Calling}} \\
    \midrule
    ARI &
      \makecell{$0.728$\\$\pm 0.221$} & \makecell{$0.883$\\$\pm 0.142$} & \makecell{$0.855$\\$\pm 0.152$} \\
    NMI &
      \makecell{$0.864$\\$\pm 0.075$} & \makecell{$0.919$\\$\pm 0.056$} & \makecell{$0.906$\\$\pm 0.063$} \\
    Homogeneity &
      \makecell{$0.973$\\$\pm 0.028$} & \makecell{$0.937$\\$\pm 0.051$} & \makecell{$0.931$\\$\pm 0.063$} \\
    Completeness &
      \makecell{$0.786$\\$\pm 0.119$} & \makecell{$0.909$\\$\pm 0.090$} & \makecell{$0.890$\\$\pm 0.100$} \\
    Average \# of clusters per run &
      $22.3$ & $17.0$ & $17.2$ \\
    Wall-clock time (s) &
      \makecell{$53.0$\\$\pm 14.5$} & \makecell{$219.4$\\$\pm 44.3$} & \makecell{$198.4$\\$\pm 59.4$} \\
    Average reflection iterations &
      $-$ & $3.00$ & $2.82 \pm 0.48$ \\
    Token usage (thousands) &
      \makecell{$34.1$\\$\pm 5.1$} & \makecell{$161.8$\\$\pm 29.1$} & \makecell{$148.4$\\$\pm 42.4$} \\
    Average Cost (USD) &
      $0.005$ & $0.026$ & $0.024$ \\
    \bottomrule
  \end{tabularx}
  \endgroup
\end{table}

\subsubsection{Experiment 2: Reasoning and classification}
\begin{figure*}[htbp]
  \centering
  \includegraphics[width=\textwidth]{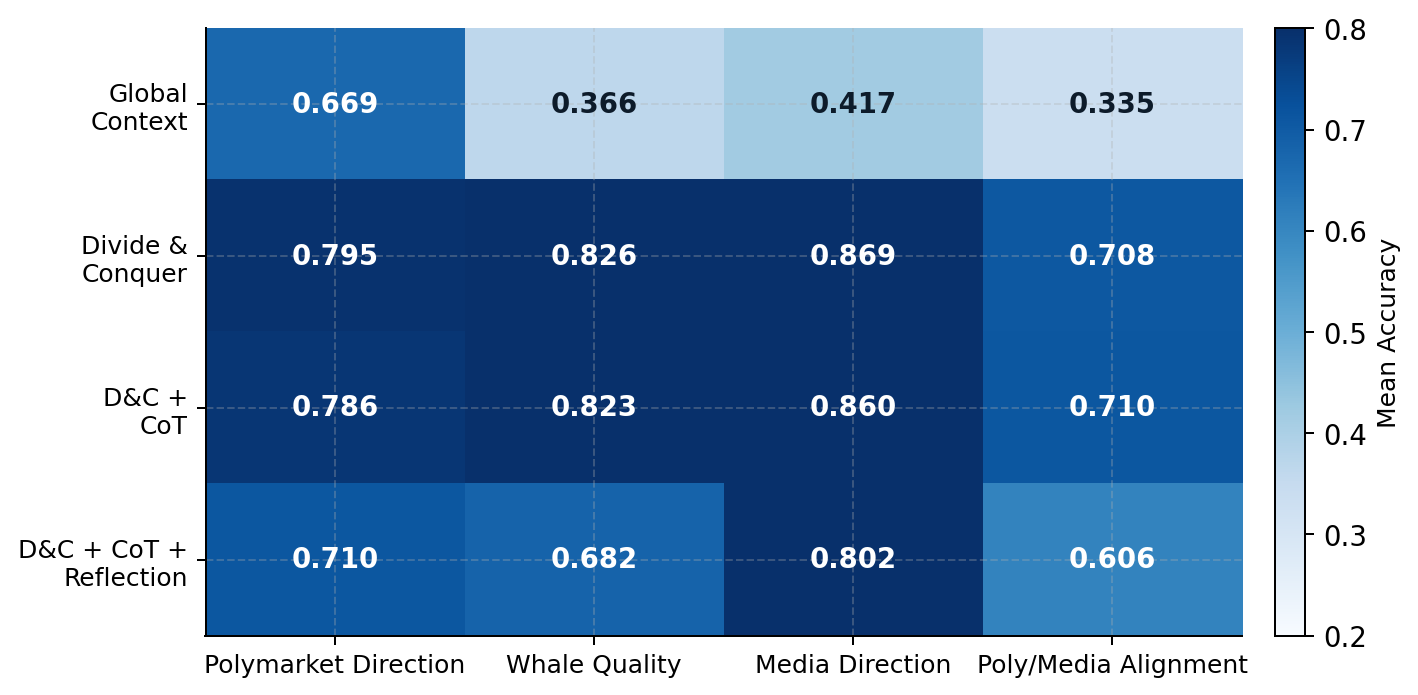}
  \caption{Experiment~2 mean classification accuracy by track and field.}
  \label{fig:exp2_heatmap}
\end{figure*}

Table~\ref{tab:exp2_results} details the performance of the Analysis Agent across the progressive reasoning harnesses. Figure~\ref{fig:exp2_heatmap} visualizes the mean accuracy heatmap across the primary evaluation dimensions (Polymarket direction, whale quality, media direction, Polymarket--media alignment). The \textit{Global Context} approach exhibits pronounced degradation in analytical synthesis, achieving an overall classification accuracy of only $0.447 \pm 0.078$. This indicates severe vulnerability to long-context noise and cross-cluster interference, characterized by attention dilution. Notably, accuracy on Whale account quality collapses to $0.366 \pm 0.066$, demonstrating the model's inability to reliably attend to granular, per-cluster metadata when all data is bundled simultaneously.

\begin{figure*}[htbp]
  \centering
  \includegraphics[width=\textwidth]{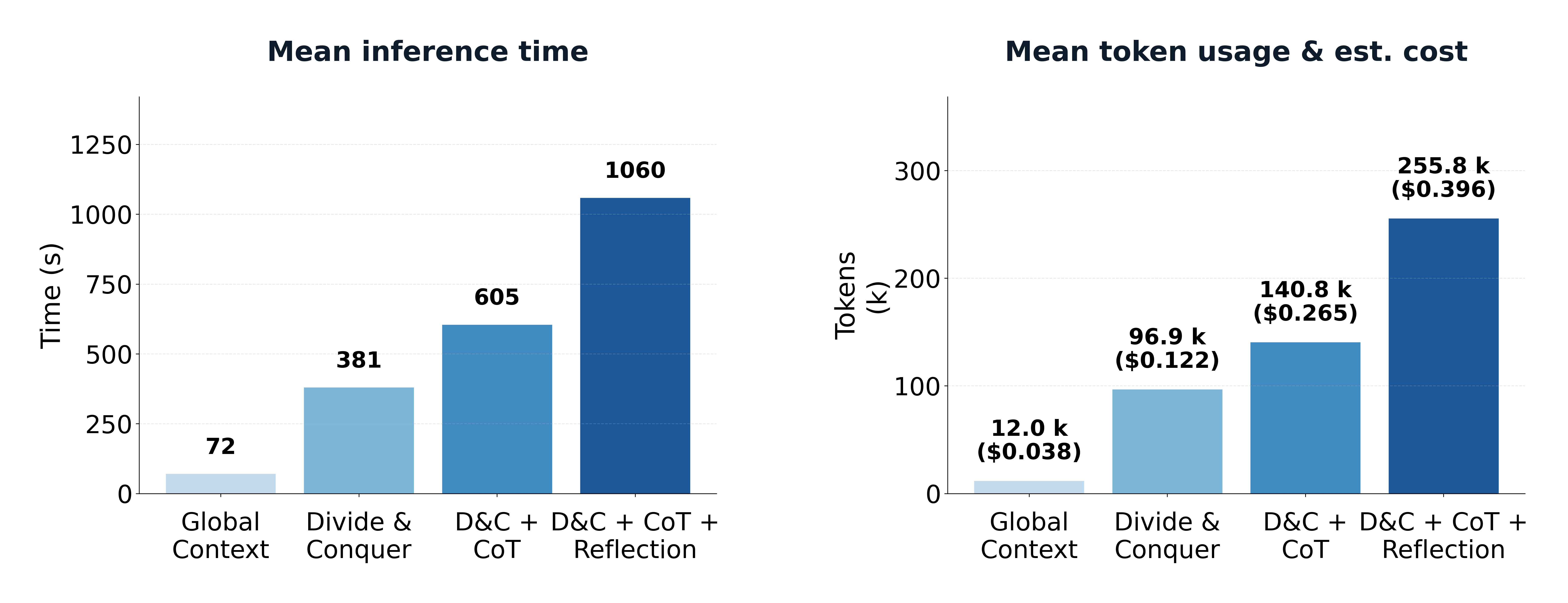}
  \caption{Experiment~2 mean inference time (left) and mean token usage with estimated cost (right) by track. Bars report track means aligned with Table~\ref{tab:exp2_results}.}
  \label{fig:exp2_time_tokens_cost}
\end{figure*}

Implementing a structural \textit{Divide-and-Conquer (D\&C)} strategy effectively mitigates this bottleneck, significantly raising overall accuracy to $0.799 \pm 0.046$ and improving generative consistency (mean pairwise cosine similarity increasing from $0.610$ to $0.696$). The explicit addition of \textit{Chain of Thought (CoT)} maintains this accuracy threshold ($0.795 \pm 0.036$) while enforcing transparent logical steps, though it offers negligible accuracy gains over D\&C alone. Notably, the \textit{D\&C + CoT + Reflection} track reveals a detrimental over-correction dynamic within this specific context; the secondary verification loop diminishes overall accuracy to $0.700 \pm 0.055$ while maximizing latency ($1059.7$ seconds) and computational overhead (\$0.396 USD). See Figure~\ref{fig:exp2_time_tokens_cost} for time and computational costs in detail. These findings establish that while architectural partitioning (D\&C) is strictly necessary for multi-dimensional OSINT alignment, unconstrained multi-agent reflection during the terminal reasoning phase can introduce adversarial logical drift.

\begin{table}[htbp]
  \centering
  \caption{Experiment~2 analysis results (mean $\pm$ std over 50 runs, 5 days x 10 runs/day).}
  \label{tab:exp2_results}
  \begingroup
  \small
  \setlength{\tabcolsep}{7pt}
  \renewcommand{\arraystretch}{1.85}
  \setlength{\extrarowheight}{4pt}
  \renewcommand{\tabularxcolumn}[1]{m{#1}}
  \begin{tabularx}{\textwidth}{@{} >{\raggedright\arraybackslash}m{3.85cm} *{4}{>{\centering\arraybackslash}X} @{}}
    \toprule
    \textbf{Metric} &
    \makecell{\textbf{Global}\\\textbf{Context}} &
    \makecell{\textbf{Divide}\\\textbf{\& Conquer}} &
    \makecell{\textbf{D\&C}\\\textbf{+ CoT}} &
    \makecell{\textbf{D\&C + CoT}\\\textbf{+ Reflection}} \\
    \midrule
    Overall classification accuracy &
      \makecell{$0.447$\\$\pm 0.078$} & \makecell{$0.799$\\$\pm 0.046$} & \makecell{$0.795$\\$\pm 0.036$} & \makecell{$0.700$\\$\pm 0.055$} \\
    Polymarket direction &
      \makecell{$0.669$\\$\pm 0.098$} & \makecell{$0.795$\\$\pm 0.076$} & \makecell{$0.786$\\$\pm 0.067$} & \makecell{$0.710$\\$\pm 0.101$} \\
    Whale account quality &
      \makecell{$0.366$\\$\pm 0.066$} & \makecell{$0.826$\\$\pm 0.055$} & \makecell{$0.823$\\$\pm 0.035$} & \makecell{$0.682$\\$\pm 0.099$} \\
    Media direction &
      \makecell{$0.417$\\$\pm 0.173$} & \makecell{$0.869$\\$\pm 0.070$} & \makecell{$0.860$\\$\pm 0.065$} & \makecell{$0.802$\\$\pm 0.072$} \\
    Polymarket--media alignment &
      \makecell{$0.335$\\$\pm 0.115$} & \makecell{$0.708$\\$\pm 0.102$} & \makecell{$0.710$\\$\pm 0.071$} & \makecell{$0.606$\\$\pm 0.098$} \\
    Mean pairwise cosine similarity (\textit{sentence-transformers}) &
      $0.610$ & $0.696$ & $0.696$ & $0.683$ \\
    Wall-clock time (s) &
      \makecell{$71.6$\\$\pm 5.9$} & \makecell{$380.7$\\$\pm 19.5$} & \makecell{$605.2$\\$\pm 43.6$} & \makecell{$1059.7$\\$\pm 66.3$} \\
    Token usage (thousands) &
      \makecell{$12.0$\\$\pm 0.7$} & \makecell{$97.0$\\$\pm 1.7$} & \makecell{$140.8$\\$\pm 2.0$} & \makecell{$255.8$\\$\pm 11.5$} \\
    Average Cost (USD) &
      $0.038$ & $0.122$ & $0.265$ & $0.396$ \\
    \bottomrule
  \end{tabularx}
  \endgroup
\end{table}

\subsection{Interpretation and Analysis}
\textbf{Clustering Dynamics and the Variance of Tool-Calling.} 
The \textit{Reflection Loop} track emerges as the Pareto-optimal solution for the clustering phase, delivering the highest accuracy alongside the lowest ARI variance. By establishing a verification boundary, the model successfully identifies over-split groupings, merges them, and validates cluster coherence without relying on external knowledge. 

Conversely, the addition of search grounding (\textit{Reflection + Tool-Calling}) introduces variance rather than net accuracy. The utility of external retrieval is highly circumstantial. As observed in the daily breakdown, Tool-Calling achieved the highest single-day ARI ($0.946$ on Day 2), suggesting that external search provides critical value when rapid, novel events are underrepresented in the LLM's static training data. However, the non-deterministic nature of search results creates a substantial downside. Poorly formed queries or off-topic retrieval results actively mislead the reflection agent, causing it to produce inferior clusters compared to the internal reflection baseline. This instability is reflected in the increased ARI standard deviation ($0.152$) and token variance, making system costs unpredictable for continuous deployment.

\textbf{Efficacy of Reasoning Paradigms.} 
The transition from \textit{Global Context} to \textit{Divide \& Conquer} represents the single largest step-change in reasoning performance, isolating information flows and preventing cross-topic contamination. Once this decomposition is implemented, the marginal utility of additional scaffolding diminishes rapidly. The \textit{D\&C + CoT} track yields nearly identical accuracy and semantic cosine similarity ($0.696$) compared to the standard D\&C track. This suggests that when a prompt is already isolated to a single cluster, explicit CoT constraints do not alter the foundational reasoning trajectory, yet they substantially increase token consumption and latency.

Furthermore, the deployment of a reflection loop at the reasoning stage actively degrades performance. When the primary D\&C agent produces highly accurate baseline outputs (approaching $80\%$), the secondary reflection agent exhibits a higher "break" rate than "fix" rate. This degradation is most acute in the evaluation of structured data; accuracy for Whale Quality dropped by $14.1$ percentage points (from $0.823$ to $0.682$). The reflection agent tends to reinterpret hard, structured on-chain data (e.g., PnL, trade counts) through a narrative lens, resulting in a severe loss of analytical precision.

\textbf{Systematic Bias in Polymarket--Media Alignment.} 
Across all architectural configurations, identifying the exact narrative alignment between Polymarket and media (Perspective Mismatch) proved to be the most challenging task, consistently scoring $10$ to $20$ percentage points below other dimensions. Analysis of the error distribution indicates that these failures are driven by stable, systematic bias rather than random hallucination. The variance for this field remains extremely low, meaning the model repeatedly makes the same localized errors.

Specifically, the model exhibits a strong attractor bias toward generating a "consensus" label. In numerous instances where the categorical outputs for Polymarket direction and media direction were diametrically opposed, or where media coverage was entirely absent, the model still synthesized a final conclusion of "consensus," actively overriding its own prior deductions and the explicit prompt rubric. This suggests that for deep narrative reasoning, standard LLMs default to agreeable, convergent summaries unless strictly constrained.

\textbf{Recommendations for Industrial Application.}
Based on these empirical findings, deploying this architecture in a live and application environment requires a selective approach to harness engineering:

\begin{itemize}
    \item \textbf{Clustering Strategy:} Deploy the internal \textit{Reflection Loop} without tool-calling for standard continuous operation to guarantee stability and cost-efficiency. Tool-calling should be strictly quarantined and triggered only by specific novelty thresholds (e.g., high-volume alerts with zero existing semantic matches in the database) and must include a stringent search-result filtering mechanism prior to ingestion.
    \item \textbf{Reasoning Strategy:} The standard \textit{Divide \& Conquer} architecture is sufficient. CoT and terminal reflection loops should be deprecated to optimize inference speed and API costs, as they offer no measurable upside for isolated cluster analysis.
    \item \textbf{Alignment Mitigation:} To counteract the model's systematic bias toward consensus, narrative reasoning must be mechanically decoupled from categorical evaluation. It is recommended to inject a deterministic, lightweight validator pass into the pipeline. This module should mathematically enforce rubric logic based purely on the independently committed categorical fields (Polymarket direction vs. Media direction) without re-reading the raw text signals, thereby preventing narrative-driven logical drift.
\end{itemize}

\section{Challenges and Future Research Opportunities}

While the proposed architecture demonstrates strong empirical performance, deploying Large Language Model (LLM)-based autonomous agents within dynamic prediction markets introduces structural and operational challenges that warrant further investigation.

\textbf{Generalization Across Evolving Market Dynamics:} Although strict temporal isolation protocols were enforced to prevent data leakage during evaluation, the underlying data-generating processes in prediction markets are inherently non-stationary. Harness architectures and prompt structures optimized for a specific temporal window may experience performance decay when exposed to novel geopolitical or macroeconomic paradigms. Future research should explore the integration of autonomous meta-agents designed to continuously evaluate and dynamically update prompt configurations, thereby maintaining alignment with shifting market narratives.

\textbf{Limits of Human Annotation:} System evaluation reliant on supervised ground-truth data is inherently bounded by the limits of human cognitive and domain expertise. Professional traders are subject to cognitive biases, and highly ambiguous market signals frequently generate significant inter-annotator disagreement. Subsequent iterations of this framework should investigate unsupervised or weakly supervised learning paradigms to autonomously extract latent predictive signals directly from multimodal data streams, effectively reducing the dependency on manual labeling.

\textbf{Dynamic Agentic Architectures:} The current framework relies on a static, rule-based harness engineering topology. As the reasoning capabilities of foundational models advance, rigidly predefined pipelines may constrain optimal task execution. Future architectures could incorporate autonomous planning modules, permitting the system to dynamically configure its retrieval and reasoning pathways based on real-time task complexity. A primary technical challenge remains balancing this dynamic flexibility with the strict execution stability and low-variance outputs.

\textbf{Latency and Computational Overhead:} As demonstrated by the empirical results, advanced reasoning scaffolds increase token utilization and inference latency. In live environments, delayed signal execution can rapidly invalidate the underlying market opportunity. Future research must address computational optimization, potentially through knowledge distillation or the deployment of specialized, smaller-parameter models (SLMs) tailored for specific sub-tasks, to better optimize the trade-off between reasoning depth and execution speed.

\section{Conclusion}
In this paper, we presented \textbf{PolyGnosis 2.0}, an advanced multi-agent framework designed to automate the fusion of \textit{Polymarket} anomalies with global \textit{OSINT} data for high-fidelity \textit{Insight Extraction}. By rigorously evaluating various \textit{Agentic Harness Engineering} configurations, we demonstrated that targeted architectural scaffolding can substantially enhance \textit{LLM Reasoning} capabilities without requiring foundational model scaling. Our empirical results establish that internal reflection loops optimize structural clustering, while a divide-and-conquer paradigm is strictly necessary to mitigate cross-topic contamination during complex narrative synthesis. 

However, we also identified critical operational limitations: unconstrained external tool-calling introduces system variance, and terminal reflection during logical synthesis frequently induces adversarial logical drift alongside escalating computational costs. Furthermore, the persistent consensus bias observed in LLM outputs highlights the necessity of implementing deterministic validators for final insight alignment. Ultimately, PolyGnosis 2.0 provides a scalable, empirically validated blueprint for deploying autonomous agents in prediction markets, successfully balancing analytical depth with the strict latency and cost constraints demanded by real-world trading environments.

\bibliographystyle{unsrtnat}
\bibliography{ref}

@article{leetaru2013gdelt,
  title={Gdelt: Global data on events, location, and tone, 1979--2012},
  author={Leetaru, Kalev and Schrodt, Philip A},
  booktitle={ISA annual convention},
  volume={2},
  number={4},
  pages={1--49},
  year={2013},
  organization={Citeseer}
}

@article{yang2025llm,
  title={LLM-as-a-Prophet: Understanding Predictive Intelligence with Prophet Arena},
  author={Yang, Qingchuan and Mahns, Simon and Li, Sida and Gu, Anri and Wu, Jibang and Xu, Haifeng},
  journal={arXiv preprint arXiv:2510.17638},
  year={2025}
}

@article{turtel2025outcome,
  title={Outcome-based Reinforcement Learning to Predict the Future},
  author={Turtel, Benjamin and Franklin, Danny and Skotheim, Kris and Hewitt, Luke and Schoenegger, Philipp},
  journal={arXiv preprint arXiv:2505.17989},
  year={2025}
}

@article{turtel2025llms,
  title={Llms can teach themselves to better predict the future},
  author={Turtel, Benjamin and Franklin, Danny and Schoenegger, Philipp},
  journal={arXiv preprint arXiv:2502.05253},
  year={2025}
}

@article{myers2025talking,
  title={Talking to gdelt through knowledge graphs},
  author={Myers, Audun and Vargas, Max and Aksoy, Sinan G and Joslyn, Cliff and Wilson, Benjamin and Burke, Lee and Grimes, Tom},
  journal={arXiv preprint arXiv:2503.07584},
  year={2025}
}

@inproceedings{palmieri2025framework,
  title={A framework for embedding generative and agentic AI in Open Source Intelligence},
  author={Palmieri, Eduardo Almeida and Ghanem, Mohamed Chahine and Sowinski-Mydlarz, Viktor and Dunsin, Dipo},
  booktitle={2025 7th International Conference on Blockchain Computing and Applications (BCCA)},
  pages={838--844},
  year={2025},
  organization={IEEE}
}

@article{berzinji2024utilisation,
  title={Utilisation of large language models (LLMs) in OSINT-based cyberterrorism detection on social media},
  author={Berzinji, Ala and Abdalmajid, Mazyar Farhad},
  journal={International Journal of Cyber Criminology},
  volume={18},
  number={1},
  pages={210--223},
  year={2024}
}

@article{chen2026cyberthreat,
  title={CyberThreat-Eval: Can Large Language Models Automate Real-World Threat Research?},
  author={Chen, Xiangsen and Feng, Xuan and Chen, Shuo and Maitre, Matthieu and Rakshit, Sudipto and Duvieilh, Diana and Picone, Ashley and Tang, Nan},
  journal={arXiv preprint arXiv:2603.09452},
  year={2026}
}

@article{comanici2025gemini,
  title={Gemini 2.5: Pushing the frontier with advanced reasoning, multimodality, long context, and next generation agentic capabilities},
  author={Comanici, Gheorghe and Bieber, Eric and Schaekermann, Mike and Pasupat, Ice and Sachdeva, Noveen and Dhillon, Inderjit and Blistein, Marcel and Ram, Ori and Zhang, Dan and Rosen, Evan and others},
  journal={arXiv preprint arXiv:2507.06261},
  year={2025}
}

@inproceedings{wu2024autogen,
  title={Autogen: Enabling next-gen LLM applications via multi-agent conversations},
  author={Wu, Qingyun and Bansal, Gagan and Zhang, Jieyu and Wu, Yiran and Li, Beibin and Zhu, Erkang and Jiang, Li and Zhang, Xiaoyun and Zhang, Shaokun and Liu, Jiale and others},
  booktitle={First conference on language modeling},
  year={2024}
}

@article{khattab2023dspy,
  title={Dspy: Compiling declarative language model calls into self-improving pipelines},
  author={Khattab, Omar and Singhvi, Arnav and Maheshwari, Paridhi and Zhang, Zhiyuan and Santhanam, Keshav and Vardhamanan, Sri and Haq, Saiful and Sharma, Ashutosh and Joshi, Thomas T and Moazam, Hanna and others},
  journal={arXiv preprint arXiv:2310.03714},
  year={2023}
}

@inproceedings{hong2023metagpt,
  title={MetaGPT: Meta programming for a multi-agent collaborative framework},
  author={Hong, Sirui and Zhuge, Mingchen and Chen, Jonathan and Zheng, Xiawu and Cheng, Yuheng and Wang, Jinlin and Zhang, Ceyao and Wang, Zili and Yau, Steven Ka Shing and Lin, Zijuan and others},
  booktitle={The twelfth international conference on learning representations},
  year={2023}
}

@article{wei2022chain,
  title={Chain-of-thought prompting elicits reasoning in large language models},
  author={Wei, Jason and Wang, Xuezhi and Schuurmans, Dale and Bosma, Maarten and Xia, Fei and Chi, Ed and Le, Quoc V and Zhou, Denny and others},
  journal={Advances in neural information processing systems},
  volume={35},
  pages={24824--24837},
  year={2022}
}

@inproceedings{yao2022react,
  title={React: Synergizing reasoning and acting in language models},
  author={Yao, Shunyu and Zhao, Jeffrey and Yu, Dian and Du, Nan and Shafran, Izhak and Narasimhan, Karthik R and Cao, Yuan},
  booktitle={The eleventh international conference on learning representations},
  year={2022}
}

@article{shinn2023reflexion,
  title={Reflexion: Language agents with verbal reinforcement learning},
  author={Shinn, Noah and Cassano, Federico and Gopinath, Ashwin and Narasimhan, Karthik and Yao, Shunyu},
  journal={Advances in neural information processing systems},
  volume={36},
  pages={8634--8652},
  year={2023}
}

@article{schick2023toolformer,
  title={Toolformer: Language models can teach themselves to use tools},
  author={Schick, Timo and Dwivedi-Yu, Jane and Dess{\`\i}, Roberto and Raileanu, Roberta and Lomeli, Maria and Hambro, Eric and Zettlemoyer, Luke and Cancedda, Nicola and Scialom, Thomas},
  journal={Advances in neural information processing systems},
  volume={36},
  pages={68539--68551},
  year={2023}
}

\appendix
\appendix
\begingroup\footnotesize

\section{Experiment 1: prompt excerpts}
Clustering Agent (Main and Summary) are shared across 3 tracks; Keywords Agent and Reflection are for Reflection Loop and Reflection + Tool-Calling; Reflection + Tool-Calling adds \texttt{REFLECTION\_SEARCH\_ADDON}.
\subsection{Clustering Agent (Main)}
\paragraph{\texttt{CLUSTERING\_SYSTEM}}
{\scriptsize
\begin{verbatim}
You are a senior financial analyst specializing in prediction markets.
Given a numbered list of Polymarket alerts (market title +
outcome side), cluster them by underlying real-world event or theme.

Rules:
- Each cluster = ONE coherent real-world theme
- Alerts about the same event but with opposite outcomes (Yes/No) belong in
the SAME cluster
- A cluster may contain only 1 alert
- Do NOT create catch-all clusters like "other", "misc", "mixed", or
"general". Every cluster must have a specific, meaningful theme.
- If an alert does not clearly relate to any other alert, it MUST form its
own singleton cluster with a descriptive theme derived from its market
title.
- Group alerts that share the same underlying real-world event or crisis
into ONE cluster, even if they touch different aspects (e.g. military
action, negotiations, and economic consequences of the same
conflict all belong together).
- Each cluster_id must be a short descriptive snake_case label that captures
the specific topic.
- For each alert decide whether the whale's bet signals the theme
**intensifying** or **easing**:
  • Read the market title carefully to understand what outcome the whale is
betting on.
  • Determine what a "Yes" resolution of that market means in the real world
(escalation, occurrence,
    price rise, etc.) and what a "No" resolution means (de-escalation,
absence, price fall, etc.).
  • Combine the outcome side the whale bought with the real-world meaning to
decide direction.
[... OMITTED: 5 lines ...]
    "cluster_id": "short_snake_case_id",
    "theme": "Human-readable theme label",
    "alerts": [
      {"index": 0, "direction": "intensifying"},
      {"index": 1, "direction": "easing"}
    ]
  }
]
\end{verbatim}
}

\subsection{Clustering Agent (Summary)}
\paragraph{\texttt{SUMMARY\_SYSTEM}}
{\scriptsize
\begin{verbatim}
You are a professional financial analyst writing for institutional
investors.
Given Polymarket prediction-market data for one theme, write
exactly ONE professional sentence in English summarizing the signal.

Requirements:
- Tone: institutional-grade, concise financial analyst voice
- State whether the theme is intensifying, easing, or roughly stable,
consistent with the provided score
- Name the key markets/contracts explicitly
- If some alerts point the other way, briefly note them as counter-signals
- One sentence only; aim for at most ~220 characters
- Return ONLY that sentence as plain text, no JSON, no markdown, no
surrounding quotes
\end{verbatim}
}

\subsection{Keywords Agent and Reflection}
\paragraph{\texttt{KEYWORD\_SYSTEM}}
{\scriptsize
\begin{verbatim}
You are a news research analyst specializing in GDELT (Global Database of
Events, Language and Tone).

Given a cluster of Polymarket prediction-market alerts, produce TWO
structured outputs for GDELT queries.

-----------------------------
1. actor_pairs  — for the GDELT **events** table (Actor1Name / Actor2Name
columns).
-----------------------------
Each entry is a LIST OF UPPERCASE tokens: [actor_A, actor_B].
A matching event row must contain actor_A in one actor field AND actor_B in
the other.
This AND-logic makes matches precise — it excludes unrelated events that
only mention one side.

Rules for each token:
  • Single UPPERCASE word, or a standard 2-word country/entity name (e.g.
"UNITED STATES", "SAUDI ARABIA").
  • Use: country names, leader surnames, organization names, institutional
roles (MILITARY, GOVERNMENT).
  • Do NOT write phrases, descriptions, or topic summaries.
  • Each pair captures a specific bilateral relationship relevant to the
cluster theme.

Produce 2-5 pairs. If the topic has no clear bilateral actors (e.g. pure
financial instrument or
sports competition with no geopolitical actors), use a single-token list:
e.g. ["ASSET_NAME"] or ["LEAGUE_NAME"].

-----------------------------
2. gkg_keywords — for the GDELT **GKG** table (V2Persons, V2Organizations,
V2Themes).
-----------------------------
GKG stores entity names exactly as they appear in news articles.
V2Persons field format: "Full Name,offset;Other Name,offset" — match on the
NAME portion only.

Rules:
  • Short proper names only: person full names or surnames, organization
names, location names.
  • Do NOT write descriptive phrases, topic summaries, or event
descriptions.
  • Each entry should be a noun or proper noun that would appear verbatim in
a news article about this topic.
  • 3-6 entries.

Return ONLY a JSON object (no markdown, no explanation):
{
  "actor_pairs": [["ACTOR_A", "ACTOR_B"], ["ACTOR_C", "ACTOR_D"], ...],
  "gkg_keywords": ["Name1", "Name2", ...]
}
\end{verbatim}
}

\paragraph{\texttt{REFLECTION\_SYSTEM\_BASE}}
{\scriptsize
\begin{verbatim}
You are a senior quality-assurance analyst reviewing prediction-market alert
clusters.
Your job is to find structural problems in the clustering and suggest
concrete fixes.

Perform these checks:

1. MERGE candidates
   - Clusters whose keywords or themes overlap heavily (same actors, same
event).
   - If you believe two cluster themes may be part of the same real-world
story, recommend merging.

2. SPLIT candidates
   - A cluster that mixes genuinely unrelated alerts (different events,
different regions,
     different asset classes) into one bucket.
   - Especially flag any cluster with a vague theme like "other", "misc",
"mixed", "general",
     "politics", or "market" — these are red flags.

3. RENAME candidates
   - cluster_id or theme that is too generic or misleading.
   - Suggest a more specific, descriptive label.

Return ONLY a JSON object (no markdown fences, no explanation outside the
JSON):
{
  "satisfied": true or false,
  "actions": [
    {"type": "merge", "clusters": ["cluster_a", "cluster_b"], "reason":
"..."},
    {"type": "split", "cluster": "cluster_c", "reason": "..."},
    {"type": "rename", "cluster": "cluster_d", "new_id": "...", "new_theme":
"...", "reason": "..."}
  ],
  "comments": "Brief overall assessment of clustering quality."
}

If the clustering looks good and no changes are needed, set "satisfied":
true and "actions": [].
\end{verbatim}
}

\paragraph{\texttt{REFLECTION\_SEARCH\_ADDON}}
{\scriptsize
\begin{verbatim}
You have access to Google Search — use it to check whether seemingly
unrelated cluster themes
are actually part of the same real-world news story or event.
When searching, focus on news from the relevant time period specified in the
prompt.
\end{verbatim}
}

\section{Experiment 2: prompt excerpts}
Global Context: \texttt{GLOBAL\_SYSTEM}. D\&C: \texttt{ANALYSIS\_SYSTEM}. D\&C + CoT: + \texttt{COT\_WRAPPER}, then extract template with \texttt{<COT\_TEXT>}. D\&C + CoT + Reflection: \texttt{REFLECTION\_SYSTEM}. .

\paragraph{\texttt{GLOBAL\_SYSTEM}}
{\scriptsize
\begin{verbatim}
You are a senior quantitative trader reviewing a full daily Polymarket
intelligence brief.

You will receive compressed data for ALL active clusters simultaneously.
Your job is to analyse every cluster and produce a structured classification
for each one.

For each cluster, classify:
- poly_direction: intensifying | easing | mixed
- whale_quality: top_trader | high_winrate | large_capital | mixed_quality |
no_whale_data | speculative
- media_direction: intensifying | easing | mixed | no_coverage
- poly_media_alignment: consensus | divergence | market_leads_media |
media_leads_market

Also write 1-2 sentence comments for poly, media, and alignment, plus an
importance_score (1-10) and one-sentence reason.

Return a JSON object with a single key "clusters" containing an array.
[... OMITTED: 5 lines ...]
  "whale_quality": "...",
  "media_direction": "...",
  "media_comment": "...",
  "poly_media_alignment": "...",
  "alignment_comment": "...",
  "importance_reason": "..."
}

You MUST include every cluster_id you receive. Do not skip any.
\end{verbatim}
}

\paragraph{\texttt{ANALYSIS\_SYSTEM}}
{\scriptsize
\begin{verbatim}
You are a senior quantitative trader reviewing a Polymarket intelligence
brief.

You will receive data for one cluster of related prediction markets:
  A) Polymarket alert data: market titles, outcome sides, price levels,
price changes (Delta%).
     For whale_buy clusters: trader profiles — name, win_rate, lifetime PnL,
trade count, USD position size.
  B) A one-sentence prior-stage summary of the cluster.
  C) GDELT media metrics from two sources:
       - Events table: geopolitical actor-pair events. Strong for
conflicts/diplomacy. Typically ZERO for crypto/finance — this is normal.
       - GKG (Global Knowledge Graph): news article volume and sentiment.
Works for ALL topics.

Your goal: classify three dimensions (Polymarket direction, Media direction,
alignment between them),
write a 1–2 sentence comment on each explaining the informational dynamics,
then rank importance from a senior trader's perspective.

===============================================
INPUT DATA REFERENCE
===============================================

Polymarket:
  price      = current implied probability (0–1)
  Delta          = price change % this window. Positive = more likely.
  direction  = "intensifying" (event more likely) or "easing" (less likely)
  whale_usd  = USD size of whale trade
  win_rate   = trader's historical win rate (0–1)
  pnl        = lifetime closed P&L in USD
  trades     = total trades in history

GDELT Events (geopolitical only; zero for crypto/finance):
  mentions_24h       = event records in last 24h
  total_mentions_30d = cumulative 30-day event records
  trend_24h_vs_3d    = today vs 3-day avg. >1 = accelerating
  tone_24h_vs_3d     = today tone minus 3-day avg. Negative = more hostile
[... OMITTED: 106 lines ...]
  High scores (8+) are reserved for topics where the underlying event
actually matters to markets.

  Scale:
    9–10 = Act now. High-impact event with clear edge (divergence or early
smart money).
    7–8  = Serious attention. High-impact topic with notable signal.
    5–6  = Monitor. Medium-impact or niche topic with interesting signal.
    3–4  = Low priority. Weak signals or niche topic without conviction.
    1–2  = Noise. No meaningful signal.

importance_reason — 1 sentence: name the topic's impact level and the
specific signal driving the score.

===============================================
Return ONLY a JSON object (no markdown fences):
{
  "poly_direction": "<intensifying | easing | mixed>",
  "poly_comment": "<1-2 sentences>",
  "whale_quality": "<top_trader | high_winrate | large_capital |
mixed_quality | no_whale_data>",
  "media_direction": "<intensifying | easing | mixed | no_coverage>",
  "media_comment": "<1-2 sentences>",
  "poly_media_alignment": "<consensus | divergence | market_leads_media |
media_leads_market>",
  "alignment_comment": "<1-2 sentences>"
}
\end{verbatim}
}

\paragraph{\texttt{COT\_WRAPPER}}
{\scriptsize
\begin{verbatim}
Before giving your final structured answer, think step-by-step through the
data.

STEP 1: Read through ALL the alerts carefully. Count how many are
intensifying vs easing.
         For whale clusters: note each trader's win_rate, PnL, and trade
count.
         Summarise: who are the strongest traders? What direction is capital
flowing?

STEP 2: Read the GDELT media data. Is coverage accelerating or declining?
         Compare Events (geopolitical) vs GKG (all topics). Note any
discrepancy.

STEP 3: Compare Polymarket signals with media signals. Are they telling the
same story?
         Who might know more — market participants or the press?

STEP 4: As a senior trader, how important is this topic? Consider:
         - What is the underlying event's real-world impact?
         - How clear is the signal? Is there edge or is it consensus?

Show your reasoning for each step, then provide the final JSON answer.

===================================================
IMPORTANT: After your step-by-step analysis, you MUST end with
a JSON object (no markdown fences) with exactly these fields:
{
  "poly_direction": "<intensifying | easing | mixed>",
  "poly_comment": "<1-2 sentences>",
  "whale_quality": "<top_trader | high_winrate | large_capital |
mixed_quality | no_whale_data>",
  "media_direction": "<intensifying | easing | mixed | no_coverage>",
  "media_comment": "<1-2 sentences>",
  "poly_media_alignment": "<consensus | divergence | market_leads_media |
media_leads_market>",
  "alignment_comment": "<1-2 sentences>"
}
\end{verbatim}
}

\paragraph{Extract template}
{\scriptsize
\begin{verbatim}
The following is a senior trader's chain-of-thought analysis of a Polymarket
cluster.

Extract ONLY the final structured classification from this analysis.
Return a JSON object with exactly these fields (no markdown fences):
{
  "poly_direction": "<intensifying | easing | mixed>",
  "poly_comment": "<1-2 sentences>",
  "whale_quality": "<top_trader | high_winrate | large_capital |
mixed_quality | no_whale_data>",
  "media_direction": "<intensifying | easing | mixed | no_coverage>",
  "media_comment": "<1-2 sentences>",
  "poly_media_alignment": "<consensus | divergence | market_leads_media |
media_leads_market>",
  "alignment_comment": "<1-2 sentences>",
  "importance_score": <integer 1-10>,
  "importance_reason": "<1 sentence>"
}

=== ANALYST CHAIN OF THOUGHT ===
<COT_TEXT>
\end{verbatim}}

\paragraph{\texttt{REFLECTION\_SYSTEM}}
{\scriptsize
\begin{verbatim}
You are a senior risk manager reviewing a batch of cluster analyses from a
junior analyst.
Each cluster has: chain-of-thought reasoning, and a structured
classification.

Review ALL clusters and flag ONLY the ones with issues. For each flagged
cluster, provide:
1. LOGICAL CONSISTENCY: Does the reasoning support the classifications?
2. DATA ACCURACY: Did the analyst correctly read the numbers (whale sizes,
PnL, win rates, article counts)?
3. CLASSIFICATION CORRECTNESS: Are the labels justified by the evidence?

Do NOT reveal any ground truth labels. Focus only on internal consistency
and correct data interpretation.

Return a JSON object with ONLY the flagged clusters (omit clusters that look
correct):
{
  "flagged_clusters": [
    {
      "cluster_id": "<the cluster_id>",
      "feedback": "<specific issues found>",
      "suggested_changes": {
        "poly_direction": "<corrected value or null if OK>",
        "whale_quality": "<corrected value or null if OK>",
        "media_direction": "<corrected value or null if OK>",
        "poly_media_alignment": "<corrected value or null if OK>"
      }
    }
  ],
  "overall_comment": "<brief summary: how many look correct, how many
flagged, general quality>"
}

If ALL analyses look correct, return: {"flagged_clusters": [],
"overall_comment": "All analyses are sound."}
\end{verbatim}
}

\section{Illustrative log excerpts}
Each evaluation \emph{track} produces many clusters in the same JSON shape; we show \textbf{one representative cluster} (or one reflection batch snippet) per track.

\subsection{Experiment 1}

\paragraph{Direct Baseline}
{\scriptsize
\begin{verbatim}
{
  "cluster_id": "iran_us_conflict",
  "theme": "US military action and conflict involving Iran",
  "alerts": [
    {
      "index": 9,
      "direction": "easing"
    },
    {
      "index": 21,
      "direction": "intensifying"
    }
  ]
}
\end{verbatim}}

\paragraph{Reflection Loop.}
{\scriptsize
\begin{verbatim}
{
  "satisfied": false,
  "actions": [
    {
      "type": "merge",
      "clusters": [
        "us_iran_military_escalation",
        "us_iran_conflict"
      ],
      "reason": "Cluster 0 ('us_iran_military_escalation') and cluster 12
('us_iran_conflict') both focus on US military involvement with Iran. The
keywords and actor pairs overlap significantly, suggesting they are part of
the same ongoing narrative."
    },
    {
      "type": "merge",
      "clusters": [
        "bitcoin_price_action",
        "bitcoin_price_targets"
      ],
      "reason": "Cluster 4 ('bitcoin_price_action') and cluster 14
('bitcoin_price_targets') both relate to Bitcoin price movements. The
keywords and actor pairs are very similar, indicating they represent the
same underlying market activity."
    },
[... OMITTED lines ...]
\end{verbatim}}

\subsection{Experiment 2}
\paragraph{D\&C - one cluster (\texttt{Eth Price}).}
\text Note that D\&C + CoT track has the same structured output, CoT omitted (same fields after extract).
{\scriptsize
\begin{verbatim}
{
  "poly_direction": "intensifying",
  "poly_comment": "A single whale deployed over $124k betting that ETH will
not dip to $1,800, but their abysmal 13% win rate over 4000 trades marks
this as low-quality capital, not smart money.",
  "whale_quality": "large_capital",
  "media_direction": "intensifying",
  "media_comment": "News volume on this topic has accelerated significantly
(1.76x trend) and the tone has become more negative (-0.1486 shift),
indicating a bearish narrative is building in the media.",
  "poly_media_alignment": "divergence",
  "alignment_comment": "There is a strong divergence where a large,
low-quality trader is taking a bullish stance against a rapidly intensifying
and negative media narrative, suggesting the market position is weak.",
  "importance_score": 9,
  "importance_reason": "A high-impact topic (ETH price) shows a clear
divergence, pitting a large but poorly-rated whale against an accelerating
bearish media narrative, creating a potential fading opportunity."
}
\end{verbatim}}

\paragraph{D\&C + CoT + Reflection - reflection (one flagged cluster + batch comment).}
{\scriptsize
\begin{verbatim}
{
  "flagged_clusters": [
    {
      "cluster_id": "trump_china_visit",
      "feedback": "LOGICAL CONSISTENCY: The reasoning for the
'media_direction' classification is inconsistent. The analyst correctly
notes the accelerating GDELT GKG volume (1.66x) but dismisses the
simultaneously improving tone (+0.1128). An accelerating volume of coverage
with improving sentiment constitutes a 'mixed' signal, not a purely
'intensifying' one. The analysis overweights one data point (volume) while
ignoring a contradictory one (tone), leading to an incorrect
classification.",
      "suggested_changes": {
        "poly_direction": null,
        "whale_quality": null,
        "media_direction": "mixed",
        "poly_media_alignment": null
      }
    }
  ],
  "overall_comment_batch_1": "Out of 5 analyses reviewed, 3 are sound. I
have flagged 2 clusters for specific issues. The errors involve a failure to
synthesize conflicting data points for a media signal in one case, and a
direct contradiction between the evidence presented in the reasoning and the
final classification label in the other. Overall quality is good, but
greater precision is needed when classifying nuanced signals and ensuring
consistency between reasoning and labels."
}
\end{verbatim}}

\endgroup

\end{document}